# A Generalization of the Noisy-Or Model

Sampath Srinivas*
Knowledge Systems Laboratory
Computer Science Department
Stanford University, CA 94305
srinivas@cs.stanford.edu

## Abstract

The Noisy-Or model is convenient for describing a class of uncertain relationships in Bayesian networks [Pearl 1988]. Pearl describes the Noisy-Or model for Boolean variables. Here we generalize the model to $n$ary input and output variables and to arbitrary functions other than the Boolean OR function. This generalization is a useful modeling aid for construction of Bayesian networks. We illustrate with some examples including digital circuit diagnosis and network reliability analysis.

## 1 INTRODUCTION

The Boolean Noisy-Or structure serves as a useful model for capturing non-deterministic disjunctive interactions between the causes of an effect [Pearl 1988].

The Boolean Noisy-Or can be explained as follows. Consider a Boolean OR gate with multiple inputs $U_1, U_2, \ldots, U_n$ and an output $X$. Now consider some non-determinism associated with each input defined as follows: On each input line $U_i$ a non-deterministic line failure function $\mathcal{N}_i$ is introduced (see Fig 1, considering $F$ to be a Boolean OR gate). The line failure function $\mathcal{N}_i$ takes $U_i$ as input and has a Boolean output $U_i'$. Instead of $U_i$ being connected to the OR gate we now have $U_i'$ connected to the OR gate instead.

The line failure function can be conceptualized as a non-deterministic device – there is a probability $q_i$ (called the *inhibitor* probability) that the line failure function causes a 'line failure'. When a line failure occurs on line $i$, the output of the device is $f$ (i.e., *false*) irrespective of what the input is, i.e., $U_i' = f$. When a line failure does not occur on line $i$ the device just transmits its input to its output, i.e., $U_i' = U_i$. This non-failure event occurs with probability $1 - q_i$.

---
*Also with Rockwell International Science Center, Palo Alto Laboratory, Palo Alto, CA 94301.

This overall structure induces a probability distribution $P(X|U_1, U_2 \ldots, U_n)$ which is easily computable[Pearl 1988].

When each $U_i$ is interpreted as a "cause" of the "effect" $X$, the Boolean Noisy-Or models disjunctive interaction of the causes. Each cause is "inhibited" with probability $q_i$, i.e., there is a probability $q_i$ that even when the cause $U_i$ is active, it will not affect $X$.

In a Bayesian network interpretation, each of the variables $U_i$ can be considered as a predecessor node of the variable $X$. The conditional probability distribution $P(X|U_1, U_2 \ldots, U_n)$ is computed from the probabilities $q_i$. In domains where such disjunctive interactions occur, instead of fully specifying opaque conditional probability distributions, the Noisy-Or model can be used instead. The inhibitor probabilities are few in number (one associated with each predecessor $U_i$ of $X$) and would be intuitively easier to specify because of their direct relation to the underlying mechanism of causation.

This paper generalizes the Noisy-Or model to the case where both the 'cause' variables $U_i$ and 'effect' variable $X$ need not be Boolean. Instead, they can be discrete variables with any number of states. Furthermore, the underlying deterministic function is not restricted to be the Boolean OR function, it can be any discrete function. In other words, in Fig 1, $F$ can be any discrete function.

Seen as a modeling tool, this generalization provides a framework to move from an underlying approximate deterministic model (the function $F$) to a more realistic probabilistic model (the distribution $P(X|U_1, U_2 \ldots, U_n)$) with the specification of only a few probabilistic parameters (the inhibitor probabilities).

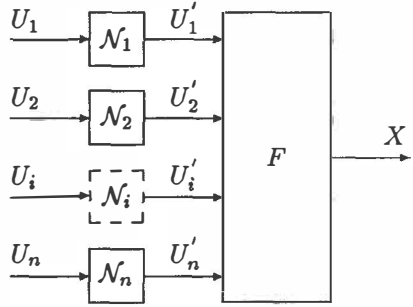

Figure 1: The generalized Noisy-Or model.

In domains where the generalized Noisy-Or is applicable, it makes the modeling task much easier when compared to the alternative of direct specification of the probabilistic model $P(X|U_1, U_2 \ldots U_n)$. In such domains, the task of creating a Bayesian network would proceed as follows:

- Variables and deterministic functions that relate them and approximate the non-deterministic behaviour of the domain are identified.

- A network is created with this information with a node for each variable, and a link from each of $U_1, U_2, \ldots, U_n$ to $X$ for each relation of form $X = F(U_1, U_2, \ldots, U_n)$. (The network is assumed to be acyclic).

- Inhibitor probabilities for each link in the network are elicited.

- The generalized Noisy-Or model is used to automatically 'lift' the network from the previous step into a fully specified Bayesian network which has the same topology as the network.

## 2  THE GENERALIZED MODEL

The generalized Noisy-Or model is illustrated in Fig 1.

Each $U_i$ is a discrete random variable. Each $U_i'$ is a discrete random variable with the same number of states as $U_i$.

We will refer to the number of states of $U_i$ and $U_i'$ as $m_i$. We will refer to the $j$th state of $U_i$ as $u_i(j)$ where $0 \leq j < m_i$. We call $j$ the index of state $u_i(j)$. We will use $u_i$ to denote "any state of $U_i$". As an example of the use of $u_i$, consider the statement, "Every state $u_i$ of $U_i$ has a unique index associated with it".

We define $I_i$ to be the function that returns the index of a state $u_i$ of $U_i$, i.e., $I_i(u_i) = j$ where $j$ is the index of state $u_i$ of variable $U_i$. We also have similarly defined quantities $u_i'(j)$, $u_i'$, $I_i'$ associated with the variable $U_i'$.

The line failure function $\mathcal{N}_i$ associates a probability value $P_i^{inh}(j)$ with every index $0 \leq j < m_i$. This quantity can be read as the inhibitor probability for the $j$th state of input $U_i$.

The line failure function can be conceptualized as a non-deterministic device that takes the value of $U_i$ as the input and outputs a value for $U_i'$. This device fails with probability $P_i^{inh}(j)$ in state $j$. When a failure in state $j$ occurs, the output of the device is $u_i'(j)$ regardless of the input. When no failure occurs, if the input is $u_i(j)$ the output is $u_i'(j)$ – this can be viewed as "passing the input through to the output" (note that the index $j$ of the output state and the input state are same in this case). The probability of no failure occuring is denoted by $P_i^{nofail}$. We see that:

$$P_i^{nofail} = 1 - \sum_{0 \leq j < m_i} P_i^{inh}(j)$$

The output $X$ is a discrete random variable with $m_x$ states. We will refer to the $j$th state of $X$ as $x(j)$ and use $x$ to refer to "any state of $X$".

$F$ (see Fig 1) can be conceptualized as a deterministic device that outputs some value $x$ of $X$ for each possible joint state $u_1', u_2', \ldots, u_n'$ of the inputs $U_1', U_2', \ldots, U_n'$. In other words $F$ is a discrete function that maps the space of joint states of $U_1' \times U_2' \times \ldots \times U_n'$ into the set of states of $X$.

We note that the model described above induces an uncertain relationship between the output $X$ and the variables $U_i$. This relationship is captured by the conditional distribution $P(X|U_1, U_2, \ldots, U_n)$.

In the next section we proceed to show how this conditional distribution is computed from the function $F$ and the inhibitor probabilities. We will use the notation $\mathbf{U}$ to denote the vector of variables $[U_1, U_2, \ldots, U_n]$. Similarly, we will use $\mathbf{u}$ to denote any joint state $[u_1, u_2, \ldots, u_n]$ of $\mathbf{U}$. $\mathbf{U}'$ and $\mathbf{u}'$ are defined similarly with respect to the variables $U_i'$. Note that $P(X|U_1, U_2, \ldots, U_n)$ abbreviates to $P(X|\mathbf{U})$.

In the special case where every inhibitor probability is zero each variable $U_i'$ always has the "same" value as $U_i$ (i.e., the state of $U_i'$ has the same index as the state of $U_i$). In this special case the variables $U_i'$ become superfluous, we could just as well remove the line failure functions and connect the each input $U_i$ directly through to $F$.

In this special case, the overall model degenerates to a deterministic function where the value of output $X$ is determined from the values of the input variables $U_i$ by the function $F$. Thus the general-ized Noisy-Or model can be viewed as starting with a deterministic model (the function $F$) and then introducing failures in the inputs, viz, the inhibitor probabilities, resulting finally in a non-deterministic model.

## 3  CHARACTERIZING $P(X|U)$

Each line failure function $\mathcal{N}_i$ defines a probability distribution $P_i(U_i'|U_i)$ relating $U_i'$ and $U_i$. From the model for $\mathcal{N}_i$ we see that the distribution $P_i$ is calculated as:

$$P_i(u_i'|u_i) = \begin{cases} P_i^{nofail} + P_i^{inh}(I_i'(u_i')) & \text{if } I_i'(u_i') = I_i(u_i) \\ P_i^{inh}(I_i'(u_i')) & \text{otherwise} \end{cases} \quad (1)$$

The equation above summarizes the following facts: if the the output $u_i'$ of $\mathcal{N}_i$ is the "same" as the input $u_i$ (i.e., the indices of both are the same), then either the device $\mathcal{N}_i$ is working normally or it has failed in the state $u_i'$. If the output $u_i'$ is not the "same" as input $u_i$, then the device has failed in state $u_i'$.

We now characterize the distribution $P(X|U)$ in terms of the inhibitor probabilities for each $U_i$ and the function $F$.

We note that:

$$P(x|\mathbf{u}) = \sum_{\mathbf{u}'} P(x|\mathbf{u}',\mathbf{u})P(\mathbf{u}'|\mathbf{u})$$

We note that once we know the state $\mathbf{u}'$ of $\mathbf{U}'$, we know the value $x$ of $X$, since $x = F(\mathbf{u}')$. In other words, $X$ is independent of $\mathbf{U}$ once $\mathbf{U}'$ is known. The above equation therefore simplifies to:

$$P(x|\mathbf{u}) = \sum_{\mathbf{u}'} P(x|\mathbf{u}')P(\mathbf{u}'|\mathbf{u})$$

We note that $P(x|\mathbf{u}') = 1$ when $x = F(\mathbf{u}')$ and $P(x|\mathbf{u}') = 0$ when $x \neq F(\mathbf{u}')$. This simplifies the defining equation to:

$$P(x|\mathbf{u}) = \sum_{\{\mathbf{u}'|x=F(\mathbf{u}')\}} P(\mathbf{u}'|\mathbf{u})$$

Now we note that the dependence of $\mathbf{u}' = [u_1, u_2, \ldots, u_n]$ on $\mathbf{u} = [u_1, u_2, \ldots, u_n]$ can be split into $n$ pairwise dependences of $u_i'$ on $u_i$. This is because the value of a variable $U_i'$ depends solely on $U_i$ and not on any other variable $U_j$ where $i \neq j$.

Thus we can simplify the equation to:

$$\begin{aligned} P(x|\mathbf{u}) &= \sum_{\{\mathbf{u}'|x=F(\mathbf{u}')\}} P(\mathbf{u}'|\mathbf{u}) \\ &= \sum_{\{\mathbf{u}'|x=F(\mathbf{u}')\}} \prod_{\mathbf{u}'} P_i(u_i'|u_i) \quad (2) \end{aligned}$$

We note that we have already defined $P(u_i'|u_i)$ in terms of the inhibitor probabilities.

The above equation is easily converted to an algorithm (described later) to generate a conditional probability table given the inhibitor probabilities and the function $F$.

### 3.1  BOOLEAN NOISY-OR AS A SPECIAL CASE

The generalized Noisy-Or collapses to be the Boolean Noisy-Or [Pearl 1988] when all the variables are Boolean[1], the function $F$ is the Boolean OR, $P_i^{inh}(0) = q_i$ and $P_i^{inh}(1) = 0$. In other words, $\mathcal{N}_i$ can fail with probability $q_i$ with the output being "false" but it cannot fail with output being "true".

Let $f_i$ and $t_i$ denote the "true" and "false" states of variable $U_i$. Similarly we have $f_x$ and $t_x$ for variable $X$. The following can be shown easily from equation 2 above:

$$\begin{aligned} P(f_x|\mathbf{u}) &= \prod_{\{i|u_i=t_i\}} q_i \\ P(t_x|\mathbf{u}) &= 1 - \prod_{\{i|u_i=t_i\}} q_i \end{aligned}$$

## 4  INTERESTING SPECIAL CASES

### 4.1  CHOICE OF A FUNCTION $F$

The generalized model described above allows the use of any discrete function $F$ relating $\mathbf{U}$ to $X$. We now suggest a particular form of $F$ that is 'compatible' with the Boolean Noisy-Or, i.e., $F$ degenerates to the Boolean OR function when the inputs and outputs are Boolean[2,3]:

$$F(\mathbf{u}') = x\left(\left\lceil (m_x - 1)\left(\frac{1}{n}\sum_i \frac{I_i'(u_i')}{(m_i - 1)}\right)\right\rceil\right)$$

In essence, this function is a weighted average – we are finding the fraction of each input's state's index over the maximum possible index of that input, averaging these fractions, scaling this quantity to the maximum index of the output, and mapping back to an actual state of the output after converting the scaled result to an integer.

---

[1] For Boolean variables we define the index of the "false" state to be 0 and the index of the "true" state to be 1.

[2] We use the syntax $\lceil \rceil$ for the Ceiling function. For a real number $x$, $\lceil x \rceil$ is the smallest integer $i$ that satisfies $i \geq x$.

[3] In the following equation, note again that $x(j)$ denotes the $j$th state of $X$.

This additive function will have the characteristic that as any input goes 'higher' it will tend to drive the output 'higher'. Further, the inputs are 'equally weighted' regardless of their arity. So, for example, a change from state 0 to state 1 in a Boolean input will have just the same effect as a change from 0 to 5 in an input with 6 states. Finally, the output is 0 if and only if all the inputs are 0.

We note that this function reduces to the Boolean OR function in the case where all inputs are Boolean and the output is Boolean.

### 4.2 CASE OF BOOLEAN OUTPUT AND nARY INPUTS

Consider the case where $X$ is a Boolean variable and the inputs $U_i$ are nary. The function $F$ is defined as in the previous section. Further, we define $P_i^{inh}(0) = q_i$ and $P_i^{inh}(j) = 0$ for $j \neq 0$. We see that we have a restricted generalization of the Boolean Noisy-Or.

This special case of nary inputs and Boolean output is interesting since it has better computational properties than the general case while being more general than the Boolean Noisy-Or (see Sec 5.2).

### 4.3 OBTAINING STRICTLY POSITIVE DISTRIBUTIONS

In some situations it is desirable for the conditional distribution of a Bayesian network node $X$ with predecessors $\mathbf{U}$ to be strictly positive, i.e., $\forall x \forall \mathbf{u} P(x|\mathbf{u}) > 0$.

For the generalized Noisy-Or model, the definition of $P(x|\mathbf{u})$ is in Equation 2. From this definition we note that the following condition is necessary to ensure a strictly positive distribution:

> For all states $x$ of $X$, the set $\{\mathbf{u}'|x = F(\mathbf{u}')\}$ is not empty. In other words, $F$ should be a function that maps $onto$ $X$.

This condition is a natural restriction – if $F$ does not satisfy this condition, the variable $X$, in effect, has superfluous states. For example, the function defined in Section 4.1 satisfies this restriction.

Assuming that the above condition is satisfied, the following condition is sufficient (though not necessary) to ensure a strictly positive distribution:

> For any $\mathbf{u}'$ and $\mathbf{u}$, $P(\mathbf{u}'|\mathbf{u}) > 0$, i.e., $\prod_{\mathbf{u}'} P_i(u_i'|u_i) > 0$.

This second condition is a stronger restriction. From Equation 1 we note that this restriction is equivalent to requiring that all inhibitor probabilities be strictly positive, i.e., that $P_i^{inh}(j) > 0$ for all $0 \leq j < m_i$.

Finally, we note that the Boolean Noisy-Or formulation of [Pearl 1988] and its generalization to nary inputs described in Section 4.2 always result in a distribution which is $not$ strictly positive since $P(t_x|\mathbf{f}) = 0$.

## 5 COMPUTING $P(X|\mathbf{U})$

We consider the complexity of generating the probabilities in the table $P(X|\mathbf{U})$.

Let $S = \prod_i m_i$ be the size of the joint state space of all the inputs $U_i$. We first note that $P_i(u_i'|u_i)$ can be computed in $\Theta(1)$ time from the inhibitor probabilities. This leads to:

$$P(\mathbf{u}'|\mathbf{u}) = \prod_i P_i(u_i'|u_i) = \Theta(n)$$

Therefore:

$$P(x|\mathbf{u}) = \sum_{\{x|x=F(\mathbf{u}')\}} P(\mathbf{u}'|\mathbf{u}) = \Theta(Sn)$$

This is because, for a given $x$ and $\mathbf{u}$ we have to traverse the entire state space of $\mathbf{U}'$ to check which $\mathbf{u}'$ satisfy $x = F(\mathbf{u}')$.

To compute the entire table we can naively compute each entry independently in which case we have:

$$P(X|\mathbf{U}) = m_x S \Theta(Sn) = \Theta(m_x n S^2)$$

However the following algorithm computes the table in $\Theta(nS^2)$:

**Begin Algorithm**
For each state $\mathbf{u}$ of $\mathbf{U}$:

- For all states $x$ of $X$ set $P(x|\mathbf{u})$ to 0.
- For each state $\mathbf{u}'$ of $\mathbf{U}'$:
  - Set $x = F(\mathbf{u}')$.
  - Increment $P(x|\mathbf{u})$ by $P(\mathbf{u}'|\mathbf{u})$.

**End Algorithm**

### 5.1 BOOLEAN NOISY-OR

In the case of the Boolean Noisy-Or, all $U_i$ and $X$ are Boolean variables. We see from Sec 3.1 that:

$$P(f_x|\mathbf{u}) = \prod_{\{i|u_i=t_i\}} q_i = \Theta(n)$$

For computing the table, we see that since $P(t_x|\mathbf{u}) = 1 - P(f_x|\mathbf{u})$, we can compute both probabilities for a particular $\mathbf{u}$ in $\Theta(n)$ time. So the time required to calculate the entire table $P(X|\mathbf{U})$ is $\Theta(Sn)$.

We see that in the case of the Boolean Noisy-Or there is a substantial saving over the general case in computing probabilities. This saving is achieved by taking into account the special characteristics of the Boolean OR function and the inhibitor probabilities when computing the distribution.

## 5.2 BOOLEAN OUTPUT AND nARY INPUTS

From an analysis similar to the previous section we note that computation of $P(X|\mathbf{U})$ takes $\Theta(Sn)$ time in this case too.

## 5.3 STORAGE COMPLEXITY

For the general case we need to store $m_i$ inhibitor probabilities per predecessor. Therefore in this case $O(nm_{max})$ storage is required where $m_{max} = \max_i(m_i)$. This contrasts with $O(m_x m_{max}^n)$ for storing the whole probability table.

For the Boolean Noisy-Or we need to store one inhibitor probability per predecessor and this is $\Theta(n)$. Using tables instead would cost $\Theta(2 \times 2^n) = \Theta(2^n)$.

In the case of nary inputs and Boolean output (as described above) one inhibitor probability per predecessor is stored. Thus storage requirement is $\Theta(n)$. Using a table would cost $O(m_{max}^n)$.

## 5.4 REDUCING COMPUTATION COMPLEXITY

In general, one could reduce the complexity of computing $P(x|\mathbf{u})$ if one could take advantage of special properties of the function $F$ to efficiently generate those $\mathbf{u}'$ that satisfy $x = F(\mathbf{u}')$ for a particular $x$.

Given a function $F$, we thus need an efficient algorithm *Invert* such that $Invert(x) = \{\mathbf{u}|x = F(\mathbf{u})\}$. By choosing $F$ carefully one can devise efficient *Invert* algorithms. However, to be useful as a modeling device, the choice of $F$ has also to be guided by the more important criterion of whether $F$ does indeed model a frequently occurring class of phenomena.

This Noisy-Or generalization has high complexity for computing probability tables from the inhibitor probabilities[4]. If the generalization is seen mostly as a useful modeling paradigm, then this complexity is not a problem, since the inhibitor probabilities can be pre-compiled into probability tables before inference takes place. Inference can be then performed with standard Bayesian network propagation algorithms.

If this generalization, however, is seen as a method of saving storage by restricting the models to a specific kind of interaction, the cost of computing the probabilities on the fly may outweigh the gains of saving space.

---

[4]However, the Boolean Noisy Or does not suffer from this problem since the special structure of the $F$ function and the fact that the inputs and outputs are Boolean reduce the complexity dramatically by a factor of $S$.

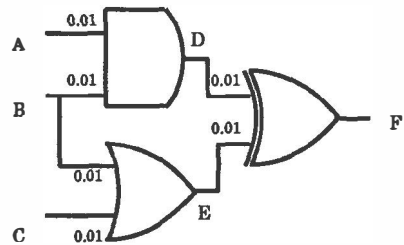

Each line has the probability of failure marked on it.

Figure 2: A digital circuit

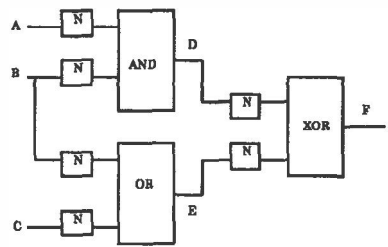

For every link the failure function $N$ has the following inhibitor probabilities (where $X$ is the predecessor variable of the link):
$$P_X^{inh}(f) = 0.01 \text{ and } P_X^{inh}(t) = 0$$

Figure 3: A generalized Noisy or model of the circuit

## 6 EXAMPLES

### 6.1 DIGITAL CIRCUIT DIAGNOSIS

The generalized Noisy-Or provides a straightforward method for doing digital circuit diagnosis. Consider the circuit in Fig 2. Let us assume that each line (i.e., wire) in the circuit has a probability of failure of 0.01 and that when a line fails, the input to the devices downstream of the line is *false*.

Each of the inputs to the devices in the circuit is now modeled with a state variable in a Noisy-Or model (see Fig 3). The function $F$ for the generalized Noisy-Or which is associated with each node is the truth table of the digital device whose output the node represents. We have an inhibitor probability of 0.01 associated with the *false* state along each link and an inhibitor probability of 0 associated with the *true* state (since the lines cannot fail in the *true* state in our fault model).

A Bayesian network is now constructed from the Noisy-Or model (see Fig 4) using the algorithm described in Section 5. Note that to complete the Bayesian network one needs the marginal distributions on the inputs to the circuit. Here we have made a choice of uniform distributions for these

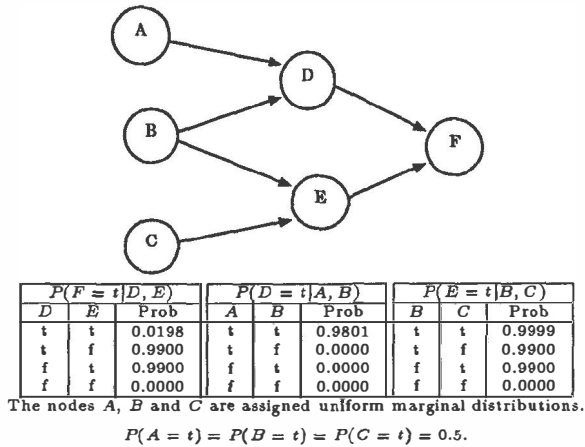

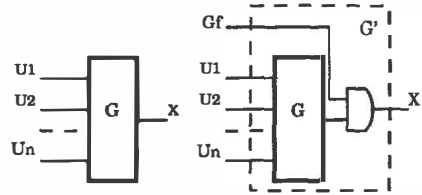

Figure 5: Modeling device failure with an 'extended' device.

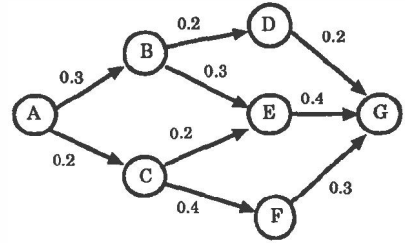

Each link has the probability of failure marked on it.

Figure 4: Bayesian network for digital circuit example.

Figure 6: A network with unreliable links.

marginals.[5]

As an example of the use of the resulting Bayesian network, consider the diagnostic question "What is the distribution of $D$ given $F$ is *false* and $B$ is *true* ?". The evidence $B = t$ and $F = f$ is declared in the Bayesian network and any standard update algorithm like the Jensen-Spiegelhalter [Jensen 1989, Lauritzen 1988] algorithm is used to yield the distribution $P(D = t|F = f, B = t) = 0.984$ and $P(D = f|F = f, B = t) = 0.016$.

Note that this example does not include a model for *device* failure – only line failures are considered. However the method can be extended easily to handle device failure by replacing every device $G$ in the circuit with the 'extended' device $G'$ as shown in Fig 5. In this figure, the input (variable) $G_f$ has a marginal distribution which reflects the probability of failure of the device. All the inhibitor probabilities on the line $G_f$ are set to 0. Note that the particular fault model illustrated here is a 'failed at *false*' model, i.e., when the device is broken, its output is false. One nice feature of the method described above is that it is incremental. If a device is added or removed from the underlying circuit a corresponding node can be added or removed from the Bayesian network – there is no need to construct a complete diagnostic model from scratch.

This method relates very well to the model based reasoning approach in this particular domain [deKleer 1987, deKleer 1989, Geffner 1987]. We describe a probabilistic approach to model-based diagnosis using Bayesian networks in detail in [Srinivas 1993b, Srinivas 1993a].

### 6.2 NETWORK CONNECTIVITY

The following example uses the Boolean Noisy-Or and the following example generalizes it to use the generalized Noisy-Or.

Consider the network shown in Fig 6. Say each link is unreliable – when the link is 'down' the link is not traversable. The reliability of each link $L$ is quantified by a probability of failure $l$ (marked on the link in the network). Now consider the question "What is the probability that a path exists from $A$ to $G$ ?".

Consider the subset of the network consisting of $A$ and its descendants (in our example, for simplicity, this is the whole network). We first associate each node with the Boolean OR as the $F$ function. Each of the link failure probabilities translates directly into the inhibitor probability for the *false* state along each link. The inhibitor probability for the *true* state is 0.

This network is now used to create a Bayesian network using the algorithm of Sec 5. The Bayesian network has the same topology as the network in Fig 6. To complete the distribution of the Bayesian network the root node $A$ has to be assigned a marginal distribution. We assign an arbitrary strictly positive distribution to the root node $A$ (since evidence is going to be declared for the root node, the actual distribution is irrelevant).

---

[5]These marginals can be seen as the distribution over the inputs provided by the environment outside the circuit. Such a distribution is not usually available. But when the distribution is *not* available, all diagnosis is *perforce* carried out with the assumption that all inputs are known. Furthermore, when all the inputs are known, it is to be noted that the answer to any diagnostic question is not affected by the actual choice of marginal as long as the marginal is any strictly positive distribution.

The answer to the question asked originally is now obtained as follows: Declare the evidence $A = t$ (and no other evidence), do evidence propagation and look at the updated belief of $G$. In this example, we get $Bel(G = t) = 0.7874$ and $Bel(G = f) = 0.2126$.[6] These beliefs are precisely the probabilities that a path exists or does not exist respectively from $A$ to $G$.

To see why, consider the case where link failures cannot happen (i.e., link failure probability is zero). Then if any variable in the network is declared to be *true* then every downstream variable to which it has some path will also be *true* due to the nature of the Boolean OR function. Once the failure probabilities are introduced, belief propagation gives us, in essence, the probability that a connected set of links existed between $A$ and $G$ forcing the OR gate at $G$ to have the output *true*.

Furthermore, it is to be noted that because belief propagation updates beliefs at *every* node, the probability of a path existing from $A$ to *any* node $X$ downstream of it is available as $Bel(X = t)$.

This method can be extended with some minor variations to answer more general questions of the form "What is the probability that there exists a path from any node in a *set* of nodes $S$ to a target node $T$ ?".

### 6.3 NETWORK CONNECTIVITY EXTENDED

Consider the exact same network as in the previous example. The question now asked is "What is the probability distribution over the *number of paths* existing from $A$ to $G$ ?".

Consider the subset of the network consisting of $A$ and its descendants. For every node $U$ we make the number of states be $n_U + 1$ where $n_U$ is the number of paths from root node $A$ to the node $U$. The states of $U$ are numbered from 0 through $n_U$. We will refer to the $i$th state of node $U$ as $u(i)$.

The number $n_U$ can be obtained for each node in the network through the following simple graph traversal algorithm:

**Begin Algorithm**

---
[6]The updated belief $Bel(X = x)$ of a variable $X$ is the conditional probability $P(X = x|E)$ where $E$ is all the available evidence.

- For root node $A$, set $n_A = 1$.[7]
- For every non root node $U$ in the graph considered in graph order (with ancestors before descendants):
  $n_U = \sum_{p \in Parents(U)} n_p$

**End Algorithm**

To build the Noisy-Or model, we now associate integer addition as the function $F$ associated with each node. For example, if $R$ and $S$ are parents of $T$ and the state of $R$ is known to be $r_2$ and the state of $S$ is known to be $s_3$, then the function maps this state of the parents to state $t_{(2+3)} = t_5$ of the child $T$.

We now set the inhibitor probabilities as follows: Say the predecessor node of some link $L$ in the graph is a node $U$. We set the inhibitor probability for state $u(0)$ to be the link failure probability $l$ and all other inhibitor probabilities to be 0. That is $P_U^{inh}(0) = l$, where $l$ is the link failure probability and $P_U^{inh}(i) = 0$ for $i = 1, 2 \ldots, n_U$.

We now construct the Bayesian network from the network described above. The marginal probability for the root node is again set arbitrarily to any strictly positive distribution since it has no effect on the result.

The answer to the question posed above is obtained by declaring the evidence $A = 1$ and then doing belief propagation to get the updated beliefs for $G$. The updated belief distribution obtained for $G$ is precisely the distribution over the number of paths from $A$ to $G$.

To see why, consider the case where there are no link failures. Then when $A$ is declared to have the value 1, the addition function at each downstream nodes counts exactly the number of paths from $A$ to itself. Once the failures are introduced the exact count becomes a distribution over the number of active paths.

In this example, we get the distribution: $Bel(G = 0) = 0.2126$, $Bel(G = 1) = 0.3466$, $Bel(G = 2) = 0.2576$, $Bel(G = 3) = 0.1326$ and $Bel(G = 4) = 0.0506$. We see that $Bel(G = 0)$ is the same probability as $Bel(G = f)$ in the previous example, viz, the probability that no path exists from $A$ to $G$.

Note that after belief updating, the distribution of number of paths from $A$ to *any* node $X$ downstream of it is available as the distribution $Bel(X)$ after belief propagation.

This method can be extended with to answer more general questions of the form "What is the distribution over the number of paths that originate

---
[7]We define the root node to have a single path to itself.

in any node in a *set* of nodes $S$ and terminate in a target node $T$ ?".

Another interesting example which can be solved using the generalized Noisy-Or is the probabilistic minimum cost path problem: Given a set of possible (positive) costs on each link of the network and a probability distribution over the costs, the problem is to determine the probability distribution over minimum cost paths between a specified pair of nodes.

The generalized Noisy-Or, in fact, can be used to solve an entire class of network problems [Srinivas 1993c]. The general approach is as in the examples above – the problem is modeled using the generalized Noisy-Or and then Bayesian propagation is used in the resulting Bayesian network to find the answer.

All the examples described above use the Noisy-Or model at *every* node in the network. However, this is not necessary. Some sections of a Bayesian network can be constructed 'conventionally', i.e., by direct elicitation of topology and input of probability tables while other sections where the Noisy-Or model is applicable, can use the Noisy-Or formalism.

## 7 IMPLEMENTATION

This generalized Noisy-Or model has been implemented in the IDEAL [Srinivas 1990] system. When creating a Noisy-Or node, the user provides the inhibitor probabilities and the deterministic function $F$.

IDEAL ensures that all implemented inference algorithms work with Bayesian networks that contain Noisy-Or nodes. This is achieved by 'compiling' the Noisy-Or information of each node into a conditional probability distribution for the node. The distribution is available for all inference algorithms to use.

### Acknowledgements

I thank Richard Fikes, Eric Horvitz, Jack Breese and Ken Fertig for invaluable discussions and suggestions.

## References

[deKleer 1987] de Kleer, J. and Williams, B. C. (1987) Diagnosing multiple faults. *Artificial Intelligence*, Volume 32, Number 1, 97–130.

[deKleer 1989] de Kleer, J. and Williams, B. C. (1989) Diagnosis with behavioral modes. Proc. of Eleventh International Joint Conference on AI, Detroit, MI. 1324–1330.

[Geffner 1987] Geffner, H. and Pearl, J. (1987) Distributed Diagnosis of Systems with Multiple Faults. In *Proceedings of the 3rd IEEE Conference on AI Applications*, Kissimmee, FL, February 1987. Also in *Readings in Model based Diagnosis*, Morgan Kauffman.

[Jensen 1989] Jensen, F. V., Lauritzen S. L. and Olesen K. G. (1989) Bayesian updating in recursive graphical models by local computations. Report R 89-15, Institute for Electronic Systems, Department of Mathematics and Computer Science, University of Aalborg, Denmark.

[Lauritzen 1988] Lauritzen, S. L. and Spiegelhalter, D. J. (1988) Local computations with probabilities on graphical structures and their applications to expert systems *J. R. Statist. Soc. B*, **50**, *No. 2*, 157–224.

[Pearl 1988] Pearl, J. (1988) *Probabilistic Reasoning in Intelligent Systems: Networks of Plausible Inference*. Morgan Kaufmann Publishers, Inc., San Mateo, Calif.

[Srinivas 1990] Srinivas, S. and Breese, J. (1990) IDEAL: A software package for analysis of influence diagrams. Proc. of 6th Conf. on Uncertainty in AI, Cambridge, MA.

[Srinivas 1993a] Srinivas, S. (1993) A probabilistic ATMS. Technical Report, Rockwell International Science Center, Palo Alto Laboratory, Palo Alto, CA.

[Srinivas 1993b] Srinivas, S. (1993) Diagnosis with behavioural modes using Bayesian networks. Technical Report, Knowledge Systems Laboratory, Computer Science Department, Stanford University. (in preparation).

[Srinivas 1993c] Srinivas, S. (1993) Using the generalized Noisy-Or to solve probabilistic network problems. Technical Report, Knowledge Systems Laboratory, Computer Science Department, Stanford University. (in preparation).